\documentclass[runningheads]{llncs}

\usepackage{eccv}

\usepackage{eccvabbrv}

\usepackage{graphicx}
\usepackage{amssymb}
\usepackage{booktabs}
\usepackage{wrapfig}
\usepackage{makecell} 
\usepackage{array} 
\usepackage{adjustbox}
\usepackage{xcolor}
\usepackage{listings}
\usepackage[accsupp]{axessibility}  

\usepackage{hyperref}

\usepackage{orcidlink}

\begin{document}

\title{EchoStyle: Unlocking High-Fidelity Video Stylization with Reverse Data Synthesis}

\titlerunning{EchoStyle}

\author{Huaqiu Li\textsuperscript{*}\inst{1,2} \and
Jiahao Wang\textsuperscript{*}\inst{2,3} \and
Sijia Cai\textsuperscript{\dag}\inst{2} \and 
Hualian Sheng\inst{2} \and
Bing Deng\inst{2} \and
Jieping Ye\inst{2} \and
Wenhan Luo\textsuperscript{\dag}\inst{1}
}

\authorrunning{H. Li et al.}

\institute{The Hong Kong University of Science and Technology \and
Alibaba Group \and
Xi'an Jiaotong University\\ 
\url{https://echostyle2026.github.io/} \\
\email{lihuaqiu2025@gmail.com}}

\maketitle

\begin{figure}[htbp] 
    \centering 
    \includegraphics[width=\textwidth]{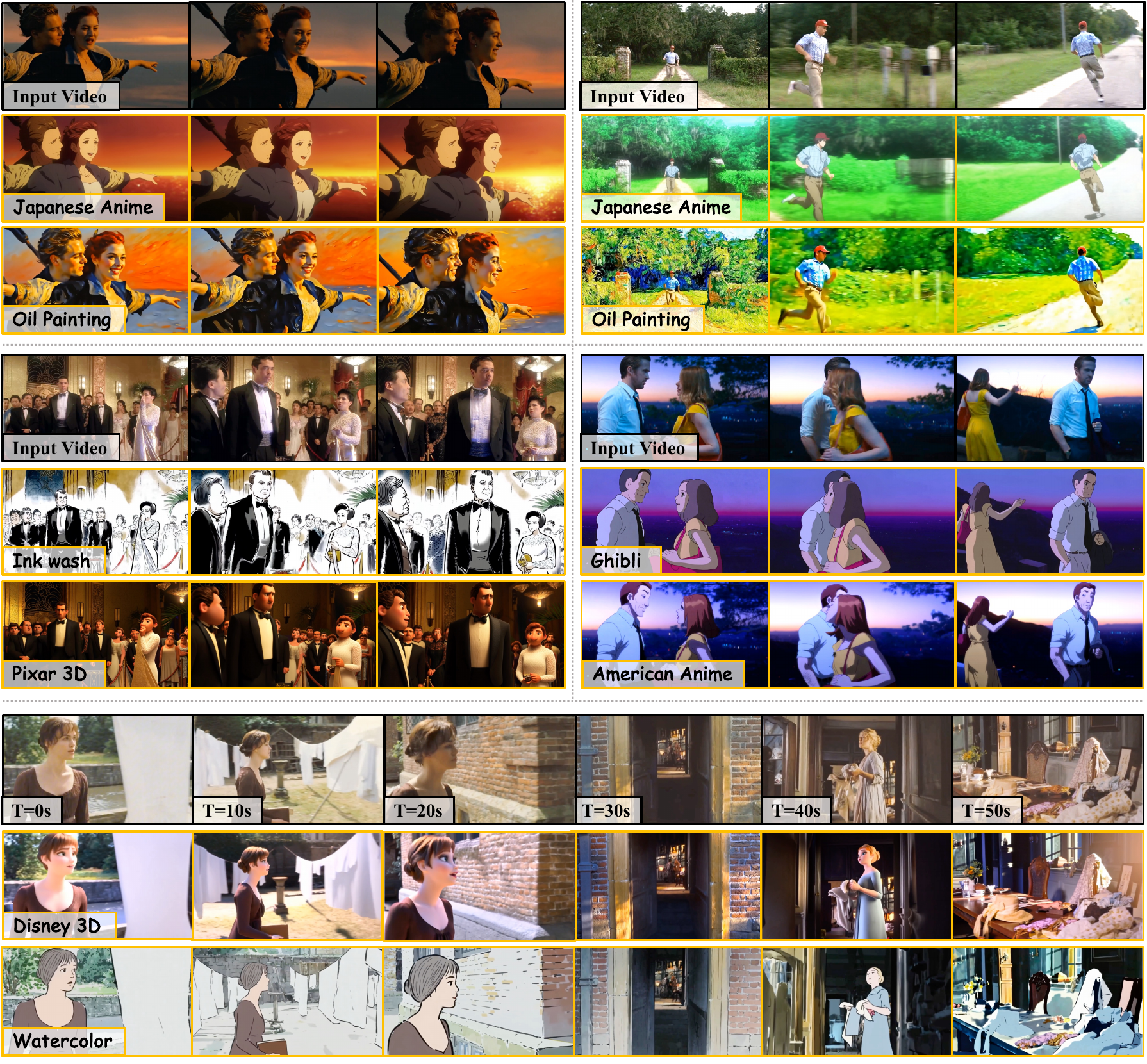} 
    \caption{EchoStyle provides a robust framework for video stylization across expansive artistic domains. By maintaining stringent temporal and motion coherence, it delivers visually compelling results that remain consistent even during long-video inference.}
    \label{fig:teaser} 
\end{figure}
\footnotetext[0]{*: Co-first Author.}
\footnotetext[1]{$\dagger$: Corresponding Author.}

\begin{abstract}
While image stylization has been studied extensively, video stylization remains a critical and largely unsolved challenge in the field of intelligent content creation. Existing methods, usually utilizing a reference image as the style prior, suffer from content leakage, data scarcity and limited adaptability to long videos, leading to suboptimal results with severe style drift and motion distortion. For these issues, we present EchoStyle, a scalable text-driven framework to achieve high-quality stylization of videos with arbitrary lengths. To start with, we construct a video-to-video architecture to appropriately re-fuse the video content and the text style. To address data scarcity, we pioneer an automatic reverse-synthesis pipeline to establish V-Style20k, a large-scale stylization dataset of 20k high-quality video pairs. To facilitate long video stylization, we devise an init-follow-mode mechanism along with a sliding-window inference strategy. Extensive experiments demonstrate EchoStyle's excellent performance across a wide range of artistic styles, even comparable to leading closed-source solutions.

\end{abstract}

\section{Introduction}
\label{sec:intro}

Recent advancements in diffusion models have sparked remarkable progress in generative modeling, triggering the evolution of controllable image and video synthesis~\cite{he2024id,stableanimator,animateany,wang2026echoshot,yang2025echomotion,li2025ld,ye2026unic,zhong2025anytalker,kong2026let,ye2026visualcot}.
Along with the development, video stylization, which aims to render a given video in a particular style, is garnering growing attention from both academia and industry for its wide-ranging applications such as artistic creation and advertisement production.

Previous works in video stylization predominantly focus on introducing a reference image as a style prior.
Conventional methods~\cite{gao2018reconet,adain,ghiasi2017exploring,shekhar2023interactive} leverage CNNs or ViTs as backbones, primarily prioritizing real-time efficiency. In recent years, there has been a significant shift towards adopting diffusion models as the foundational architecture. For instance, StyleCrafter~\cite{liu2023stylecrafter} and StyleMaster~\cite{ye2025stylemaster} adapt pre-trained text-to-video diffusion models~\cite{lvdm,wan2025wan} through post-training strategies. In parallel, training-free methods, such as AnyV2V~\cite{kuanyv2v} and FreeVis~\cite{xu2025freevis}, typically stylize keyframes via image editing models and then propagate the style through feature injection to ensure temporal consistency. Furthermore, frameworks like UniVST~\cite{song2025univst} have extended training-free capabilities to support localized video stylization.

However, these existing paradigms suffer from several critical limitations: 
\textbf{Content-style entanglement.} Since the reference image contains redundant style-irrelevant information besides style, the reliance on it often leads to unintended content leakage into the stylized videos~\cite{clipstyler,ipadapter}. Consequently, commercial closed-source video models~\cite{seedance2.0, Kling-O1} and several image stylization methods~\cite{clipstyler,nameyourstyle,suresh2024fastclipstyler} concentrate on the text-driven roadmap to achieve concept-level stylization and inherently avoid content leakage. Yet, despite their advancements, a reliable open-source text-driven video stylization paradigm remains elusive.
\textbf{Video data scarcity.} The training of video stylization requires high-quality video pairs with the same content but different styles, which are extremely scarce. Existing video stylization methods either bypass training by handcrafted components~\cite{kuanyv2v,xu2025freevis,song2025univst} or conduct training on image datasets~\cite{still,liu2023stylecrafter,ye2025stylemaster}, both of which lead to severe style degradation and instability, especially under strong video dynamics.
\textbf{Limited ability of extension.} Previous works are confined to short sequences ($\le$ 5s) and fail to provide robust, scalable designs for long-duration generation. Therefore, developing a stable and expandable framework for text-driven video stylization remains a valuable and challenging problem.

To address the aforementioned challenges, we present EchoStyle, a text-driven framework designed for high-fidelity stylization of videos with arbitrary lengths. We formalize the text-driven video stylization as video-to-video generation and propose a streamlined yet effective architecture based on the Wan2.2-I2V~\cite{wan2025wan} foundation to appropriately re-fuse the video content and the text style. To bridge the data gap caused by scarcity and inadequate quality, we devise an efficient and reusable data generation pipeline that employs reverse-synthesis to derive reference input videos from open-source stylish datasets. With such a curated pipeline, we construct V-Style20k, a pioneering large-scale video stylization dataset of 20k high-quality video pairs to drive the training process. To further achieve consistent stylization for long videos, we design a temporal recurrent generation approach, which highlights an init-follow-mode mechanism during training and a sliding-window strategy during inference. As shown in Fig.~\ref{fig:teaser}, extensive experiments demonstrate that our method achieves robust and stable stylization for both short and long videos, even in cases of complex content and intense dynamics.

We summarize our key contributions as follows:
\begin{itemize}
\item We propose a novel video-to-video framework tailored for text-driven stylization of videos with arbitrary lengths, which is naturally scalable owing to its simplicity.
\item We pioneer a fully automated data curation pipeline to construct an innovative large-scale video stylization dataset, V-Style20k, which contains 20k high-quality video pairs.
\item We incorporate an intricate init-follow-mode mechanism into the training along with a sliding-window inference strategy, enabling stable stylization for long videos spanning several minutes.
\item Comprehensive experiments demonstrate that EchoStyle achieves superior dynamic style consistency as well as fine-grained content preservation in video stylization, yielding performance comparable to closed-source models.
\end{itemize}
\section{Related Work}

\subsection{Image \& Video Stylization}
Stylization aims at applying the artistic style from the reference condition to either a given content image or video. For image stylization, early research, such as~\cite{adain,gao2018reconet}, utilizes a pre-trained network as a style encoder and a carefully designed injection module. More recently, powerful foundation and generative models, such as Stable Diffusion~\cite{sd} and FLUX~\cite{flux}, have broadened learning-based image manipulation from controllable stylization to restoration~\cite{ipadapter,deadiff,huaqiu2025interpretable,li2025ld}. Building upon this, works like InstantStyle~\cite{instantstyle}, StyleShot~\cite{gao2025styleshot}, and B-LoRA~\cite{b-lora} have proposed more granular, time-aware and layer-aware injection strategies to better disentangle the infusion of style and content features.

In the field of video stylization, conventional methods~\cite{gao2018reconet,adain,ghiasi2017exploring} predominantly rely on CNN or ViT backbones, prioritizing low-latency inference and real-time efficiency. However, recent years have witnessed a paradigm shift toward diffusion-based architectures. For instance, StyleCrafter~\cite{liu2023stylecrafter}, StyleMaster~\cite{ye2025stylemaster} and PickStyle~\cite{pickstyle} adapt pre-trained T2V models through specialized post-training strategies, while Telestyle~\cite {telestyle} tries to adapt pre-trained image generation models to the video domain. Concurrently, training-free pipelines have gained attention by leveraging non-end-to-end workflows. Approaches like AnyV2V~\cite{kuanyv2v} and FreeVis~\cite{xu2025freevis} typically stylize representative keyframes via image-editing models and subsequently propagate these stylistic attributes through feature injection to ensure temporal consistency. Furthermore, frameworks such as UniVST~\cite{song2025univst} have extended these training-free capabilities to support localized or region-aware video stylization.

\subsection{Video-to-Video Frameworks}
Existing research has extensively investigated Video-to-Video (V2V) translation frameworks. In particular, Video-P2P~\cite{videop2p} facilitates real-world video editing through cross-attention manipulation, while TokenFlow~\cite{geyer2023tokenflow} and AnyV2V~\cite{kuanyv2v} maintain temporal consistency through the propagation of cross-frame attention features. Other task-specific approaches, including Animate Anyone~\cite{animateany} and MotionCtrl~\cite{wang2024motionctrl}, leverage Denoising UNet with temporal attention and Latent Video Diffusion Model~\cite{lvdm} backbones. Furthermore, VACE~\cite{jiang2025vace} adopts a ControlNet-like mechanism to inject video guidance, achieving unified V2V generation. Despite these advancements, prevailing methodologies often emphasize generative diversity rather than input-output semantic alignment and face significant challenges in scaling to long videos. This underscores the necessity of a dedicated V2V framework specifically designed for video stylization.


\section{Method}
To facilitate high-quality and expandable text-driven video stylization, we introduce a novel video-to-video architecture to appropriately re-fuse the video content and the text style in Sec.~\ref{sec:v2v_network}. To address data scarcity, we present a robust data curation roadmap that utilizes a reverse-synthesis pipeline to construct a large-scale paired video dataset in Sec.~\ref{sec:data_pipeline}. Finally, in Sec.~\ref{sec:long_video}, we detail an intricate init-follow-mode mechanism along with a sliding-window inference strategy for long-video stylization.

\subsection{Constructing Text-Driven Video-to-Video Framework}
\label{sec:v2v_network}

\begin{figure}[t]
    \centering
    \includegraphics[width=\textwidth]{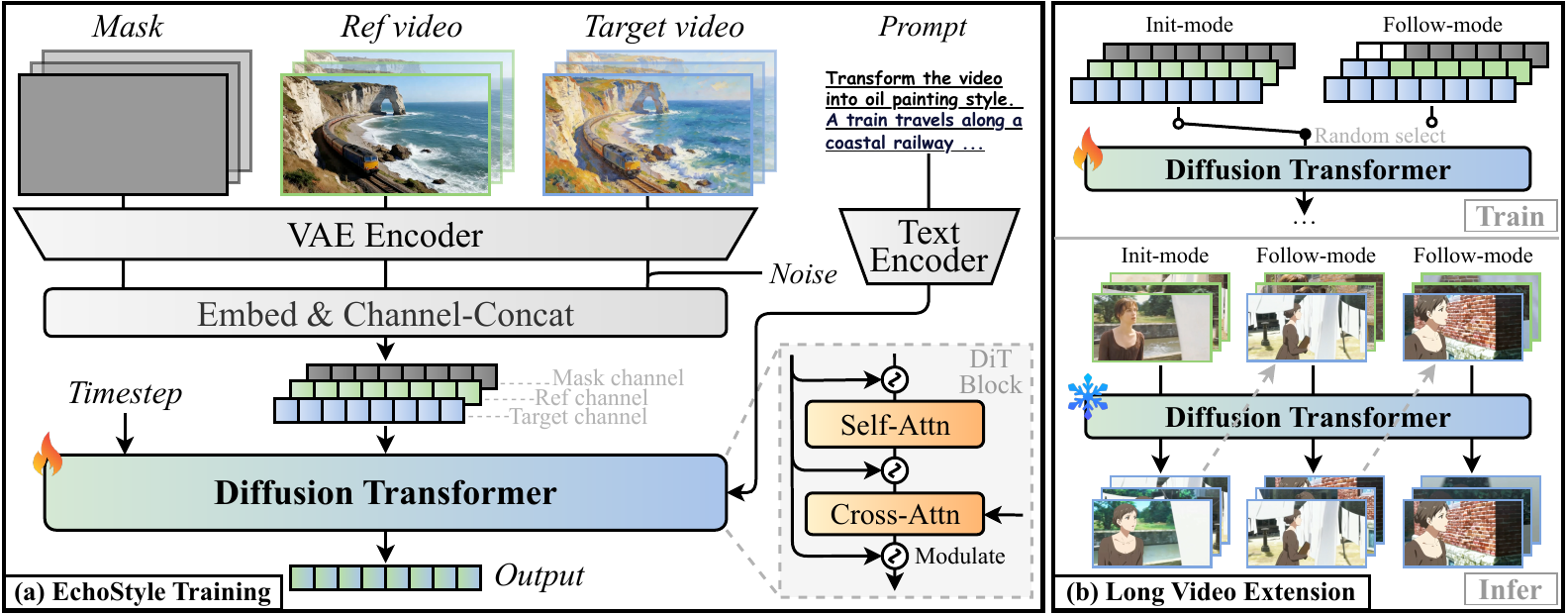} 
    \caption{Overview of the EchoStyle framework. (a) Training: We map the mask, reference video, and target video into the shared latent space, followed by visual alignment. These multi-channel embeddings are then fed into a DiT for stylized video generation, guided by textual prompts. (b) Long Video Extension: We propose an init-follow-mode training strategy. During training, these modes are randomly selected. In the inference stage, long videos are generated autoregressively, where init-mode generates the starting sequence and subsequent segments are produced via follow-mode.}
    \label{fig:pipe}
\end{figure}
Formally, we characterize text-driven video-to-video stylization through a set of input conditions: a text prompt $p$, which encompasses both target style and content descriptions, a reference video $\mathbf{R}$ to be stylized, and a mask $\mathbf{M}$ to define which frames to be modified. We expect the model to produce the output video $\mathbf{O}$ through the mapping:
\begin{equation}
    \mathbf{O} = \Phi(\mathbf{R},\mathbf{M}, p),
\end{equation}
where $\Phi(\cdot)$ denotes the generative framework, and $\mathbf{M}$ is set to 0 for every frame.

The task of video stylization primarily hinges on addressing two core challenges: achieving rigorous content alignment with the reference video and developing expandable potential. To overcome these bottlenecks, EchoStyle adopts a streamlined yet effective architecture as illustrated in Fig.~\ref{fig:pipe}(a). To specify, we use the Variational Autoencoder (VAE)~\cite{VAE} to encode the reference video $\mathbf{R}$, the target video $\mathbf{V}$ into the latent space, yielding their respective latent representations, $\mathbf{z}_R$ and $\mathbf{z}_V$.
Next, a timestep $t$ is uniformly sampled from {1, ..., $T$}. We then corrupt the target latent $\mathbf{z}_V$ with noise according to the diffusion process forward kernel, resulting in the noisy latent $\mathbf{z}_V^{(t)}$.
Concurrently, $\mathbf{M}$, which shares the same spatial and temporal dimensions as the input videos, is then reshaped to match the shape of the latent $\mathbf{z}_V^{(t)}$ and $\mathbf{z}_R$, yielding $\mathbf{z}_M$. Finally, we concatenate these three latent tensors along the channel dimension to form the unified model input $\mathbf{z}_{in}$ as:
\begin{equation}
    \mathbf{z}_{\text{in}} = \text{Concat}(\mathbf{z}_R, \mathbf{z}_V^{(t)}, \mathbf{z}_M).
    \label{equ:concat}
\end{equation}
By introducing this unified and standardized visual alignment strategy, we consolidate diverse references ($\mathbf{R},\mathbf{V},\mathbf{M}$) into a unified latent. This approach ensures high semantic coherence, spatial and temporal alignments between the input video condition and the target $\mathbf{V}$.

This concatenated tensor is then fed into the backbone for noise prediction,
which is constructed by stacking a series of Diffusion Transformer (DiT) blocks~\cite{dit}. 
The conditioning information, including the timestep $t$ and textual prompt $p$, is injected into the network through Adaptive Layer Normalization (adaLN) and Cross-Attention. The forward propagation of the DiT model can be denoted as:
\begin{equation}
    \frac{d\mathbf{z}_V^{(t)}}{dt}=\boldsymbol{\epsilon}_{\mathbf{\theta}} = \text{DiT}(\mathbf{z}_{\text{in}}, t, p),
    \label{eq:forward_pass}
\end{equation}
where $\boldsymbol{\epsilon}_{\mathbf{\theta}}$ is the prediction from our model $\mathbf{\theta}$.
We utilize the standard flow-matching loss as the training objective, which is to minimize the MSE between the prediction $\boldsymbol{\epsilon}_{\mathbf{\theta}}$ and the ground-truth $(\mathbf{z}_V^{(0)} - \boldsymbol{\epsilon})$, then update the entire model in a LoRA~\cite{lora} manner. The training objective can be denoted as:
\begin{equation}
    \mathcal{L}(\mathbf{\theta}) = \mathbb{E}_{t \sim \mathcal{U}[0, 1]} \left\| \frac{d\mathbf{z}_V^{(t)}}{dt} - (\mathbf{z}_V^{(0)} - \boldsymbol{\epsilon}) \right\|^2_2.
\end{equation}


\begin{figure}[t] 
    \centering
    \includegraphics[width=\textwidth]{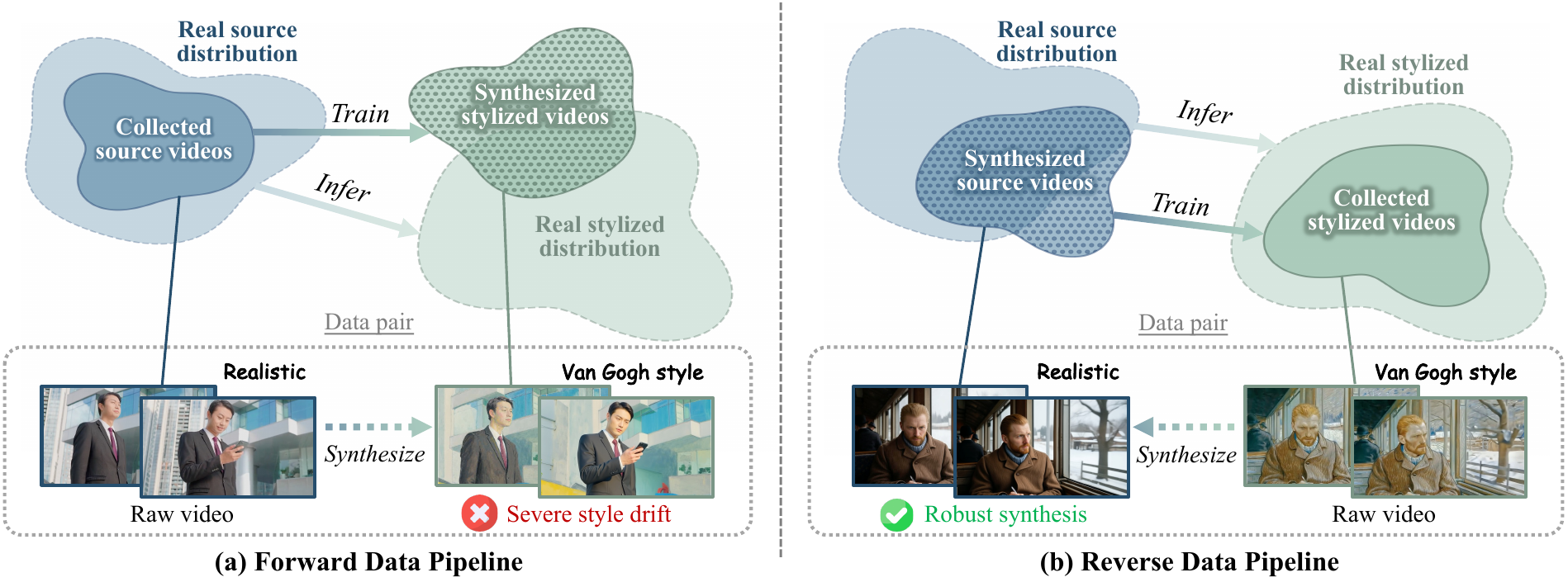} 
    \caption{Forward vs. Reverse Data Pipelines. (a) The forward pipeline suffers from significant distribution mismatch and style drift. (b) In contrast, our reverse pipeline leverages real stylized data as the source, achieving superior distribution alignment and more robust generation results.}
    \label{fig:data_compare}
\end{figure}
\subsection{Curating High-Quality Video Pairs}
\label{sec:data_pipeline}

Given the scarcity of large-scale, paired video stylization datasets, synthesizing data with existing models is a common strategy.
However, AI-synthesized stylized videos suffer from limited stylistic diversity, flickering, and style drift. As Fig.~\ref{fig:data_compare} illustrates, using such data as a training target would contaminate the objective distribution with these flaws, thereby capping the model's potential. This limitation stems from the data bias inherent in current video2video tools, where the training sets are dominated by real-world imagery. 

We effectively address the aforementioned challenges by inverting the direction of this synthesis process. To specify, we collect authentic stylized videos and synthesize their corresponding realistic counterparts to serve as source inputs. 
The primary advantage of this approach lies in defining authentic stylized videos as the target distribution, which ensures the purity of the learning objective and effectively circumvents the performance bottlenecks typically imposed by synthetic artifacts or stylistic drift. 
Furthermore, by strategically exploiting the intrinsic realism bias of pre-trained models, we achieve superior fidelity and stability in reference video construction during reverse synthesis, utilizing model bias as an advantage for high-quality data curation.

\begin{figure}[t]
    \centering
    \includegraphics[width=\textwidth]{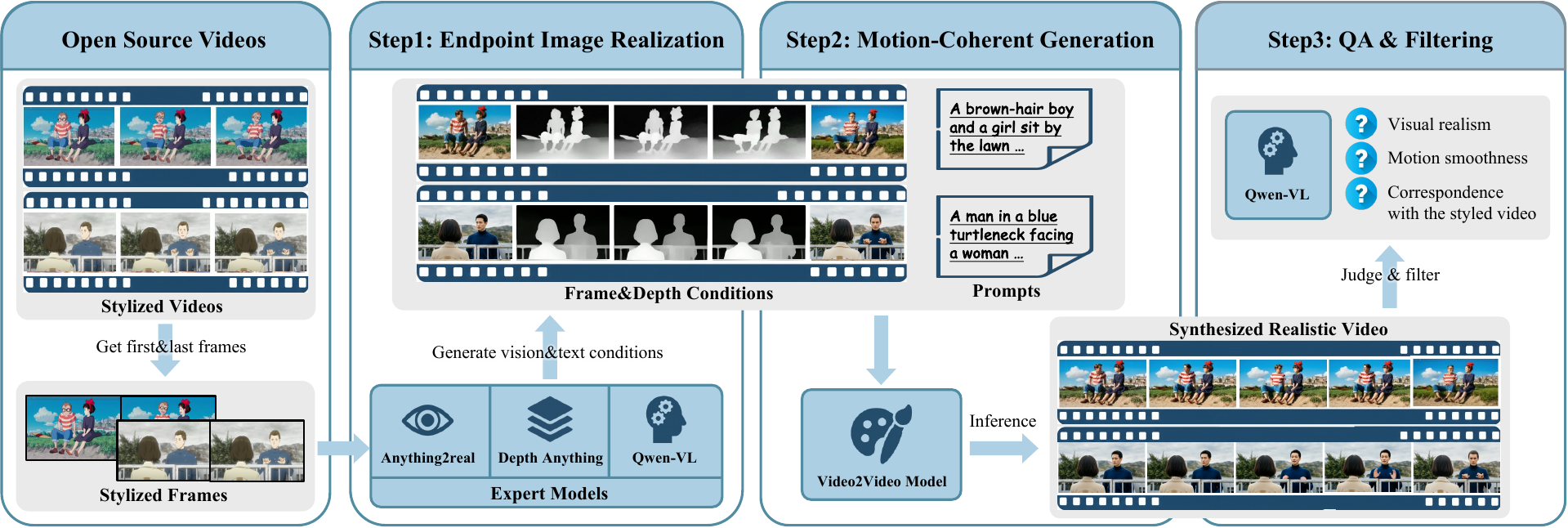} 
    \caption{Our data synthesis pipeline is organized into three stages: first, endpoint image stylization establishes the style for boundary frames; second, motion-coherent V2V generation produces semantically-aligned video sequences; and finally, automated filtering via VLMs is employed to vet video quality and discard suboptimal samples.}
    \label{fig:data_pipe}
\end{figure}
As shown in Fig.~\ref{fig:data_pipe}, our automated data collection pipeline proceeds as follows:
\begin{itemize}
\item \noindent\textbf{Stylized Video Collection.} We collect stylized video clips from open-source datasets and extract their initial and final frames.
    \item \noindent\textbf{Image Realization.}
We translate the first and final frames of the stylized videos into realistic endpoint anchors using Qwen-Image~\cite{qwenimage} as the base model and Anything-to-Real LoRA for image-to-image translation. 
\item \noindent\textbf{Motion-coherent V2V Generation.} These transformed realistic images, combined with depth priors extracted from the original stylized clip, serve as conditions for VACE~\cite{jiang2025vace}. This depth and first \& last frame-guided process ensures the generated realistic video maintains precise content coherence with the stylized video.
\item \noindent\textbf{Quality Assessment.}
To ensure data fidelity, we utilize Qwen-VL~\cite{qwenvl} to screen video pairs based on three key metrics: (i) generated visual realism, (ii) motion plausibility, and (iii) fine-grained temporal correspondence. Only pairs that meet these rigorous criteria are incorporated into the final training dataset.
\end{itemize}

\begin{figure}[t]
    \centering
    \includegraphics[width=\textwidth]{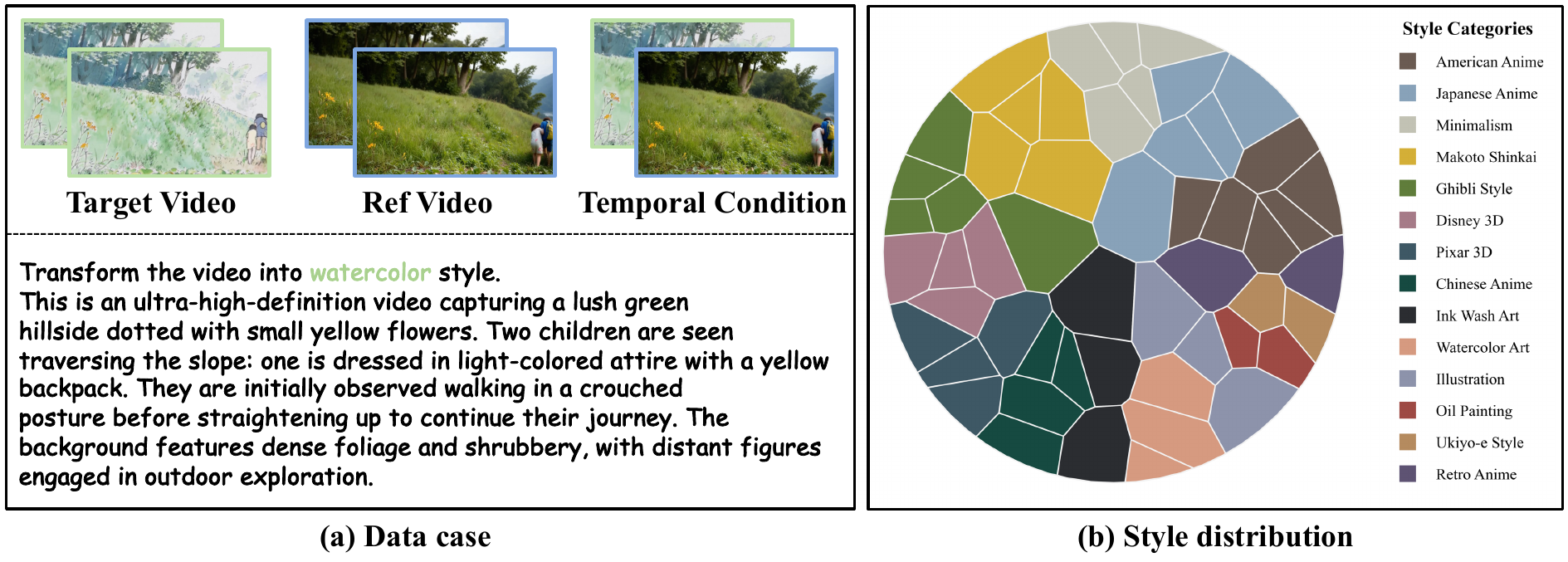}
    \caption{Representative samples and statistical characterization of V-Style20k.}
    \label{fig:dataset}
\end{figure}

\textit{V-Style20k} \textbf{Dataset.} 
Leveraging this reverse data pipeline, we establish \textit{V-Style20k}, a large-scale paired video dataset designed to facilitate research in video stylization. The dataset comprises over 20,000 high-quality video pairs at a resolution of 480×832, with each clip restricted to a duration of under 5 seconds. \textit{V-Style20k} encompasses 14 diverse styles prevalent in cinematography, animation, and fine arts. Furthermore, to support text-driven tasks, each sample is meticulously annotated with descriptive captions using the Qwen-VL~\cite{qwenvl} vision-language model.
We formally denote our dataset $D$ as:
\begin{equation}
    D = \{(p_j, \mathbf{R}_j, \mathbf{V}_j, \mathbf{T}_j)|j=1,2,\dots,N\}.
\end{equation}
Specifically, each data sample comprises a text prompt $p$ describing the intended style and content, an input reference video $\mathbf{R}$ serving as the content prior, a target stylized video $\mathbf{V}$, and a $\mathbf{T}$ serving as a temporal condition which will be thoroughly discussed in Sec.~\ref{sec:long_video}. We provide the illustration and statistics of our dataset in Fig.~\ref{fig:dataset}.

\subsection{Expanding to Long-Video Stylization}
\label{sec:long_video}




To mitigate the pervasive memory bottlenecks in long-video generation and achieve expandable stylization, we propose a sliding-window-based generation framework powered by an init-follow-mode mechanism during training, as shown in Fig.~\ref{fig:pipe}(b). Specifically, in the follow mode, we introduce a temporal reference video $\mathbf{T}$, which is constructed by replacing the initial $x$-frame window of $\mathbf{R}$ with the corresponding stylized frames from $\mathbf{V}$. During training, we utilize VAE to compress $\mathbf{T}$ into $\mathbf{z}_T$. Then we replace $\mathbf{z}_R$ in Eq.~(\ref{equ:concat}) with $\mathbf{z}_T$ to realize dual-mode training, and modify the respective frames in the mask $\mathbf{M}$ as 1 to indicate that these frames need to be preserved, which can be denoted as:
\begin{equation}
    \mathbf{z}_{\text{in}} = \text{Concat}(\mathbf{z}_T, \mathbf{z}_V^{(t)}, \mathbf{z}_M).
\end{equation}
In each training iteration, we choose $\mathbf{z}_{\text{in}}$ from two modes with a fixed probability, ensuring that the network effectively balances its capacity.

During inference, instead of relying solely on the first frame, we utilize clip-guided conditioning to reduce the sensitivity to one single frame's quality and enhance the temporal stability of the stylized output. We partition the long-form video into a sequence of shorter segments $\{\mathbf{R}_{0}, \mathbf{R}_{1}, \dots, \mathbf{R}_{n}\}$, where each consecutive pair $(\mathbf{R}_{i}, \mathbf{R}_{i+1})$ shares a temporal overlap of $x$ frames.
The initial segment $\mathbf{R}_{\mathbf{0}}$ is processed following the short-duration pipeline. 
For subsequent segments $\mathbf{R}_{i}$ ($i > 0$), the generation is conditioned on the overlapping stylized frames from the previous output $\mathbf{O}_{i-1}$. 
Formally, let $\mathbf{O}_{i-1}[-x:]$ denote the last $x$ frames of the preceding stylized segment. $\mathbf{M}_x$ means setting the first $x$ frames of $\textbf{M}$ to 1 to indicate that these frames should be modified. The inference for $\mathbf{R}_{i}$ is formulated as:
\begin{equation}
    \mathbf{O}_{i} = \Phi(\mathbf{R}_{i},\mathbf{M}_x, p_{i}, \mathbf{O}_{i-1}[-x:]),
\end{equation}
This process ensures smooth and temporally coherent generation across long video sequences. The equation can be formulated as:
\begin{equation}
\mathbf{O}_{\text{long}} = \mathbf{O}_0 \mathbin{\Vert} \sum_{i=1}^{n} \mathbf{O}_i[x:].
\end{equation}
where $\mathbin{\Vert}$ denotes the concatenation along the temporal axis. This init-follow-mode helps us to seamlessly integrate the training and inference processes in variable-length video stylization into a unified pattern, not only enhancing the structural simplicity but also significantly reducing computational costs.

\section{Experiment}
\subsection{Implementation Details}

\textbf{Data Preparation.} 
To facilitate our training process, we utilize \textit{V-Style20k} during our training process. Videos are center-cropped to a spatial resolution of $480 \times 480$ to maintain consistency. We adopt a temporal bucket strategy to handle varying sequence lengths, with frame counts ranging from 33 to 69 frames, all sampled at a frame rate of 16 FPS.

\textbf{Base Model and Components.} 
Our framework is initialized with the pre-trained weights of Wan2.2-I2V-14B, spanning both high-noise and low-noise regimes. During the fine-tuning stage, we freeze the VAE and text encoders, updating only the LoRA parameters integrated into the DiT backbone. Specifically, we inject LoRA into all \texttt{nn.Linear} layers with a rank of $r=64$ and $\alpha=32$, utilizing Gaussian initialization. Training is orchestrated via torchrun on a cluster of 16 NVIDIA A100 GPUs. We set the global seed to 42 and apply a per-GPU batch size of 1, utilizing a $seed+rank$ strategy to ensure multi-process decorrelation. The model is trained for $20,000$ steps, totaling approximately 80 hours of computation.

\textbf{Training Strategy.} 
To enhance the model's robustness and controllability, we employ a diverse sampling strategy for text and video conditioning during each iteration. For textual guidance, the model is provided with full captions (a concatenation of style instructions and content descriptions) with a probability of 0.8; otherwise, only style instructions are used. 
Regarding visual conditioning, we utilize reference video $\mathbf{R}$ with all-zero mask $\mathbf{M}$ in 80\% of the samples. For the remaining 20\%, we use the temporal reference video $\mathbf{T}$ with the modified mask $\mathbf{M}_x$, and $x$ is set to 16 during training. 
We leverage mixed-precision training, where non-LoRA weights are converted to \texttt{bfloat16} to reduce memory overhead. Optimization is performed using the AdamW optimizer with $\beta=(0.9, 0.999)$, a weight decay of $0.01$, and a constant learning rate of $2 \times 10^{-5}$.
\subsection{Comparison Experiment}
To benchmark our method against both open-source and closed-source models, we assemble an evaluation set of 50 video-instruction pairs featuring nine representative styles. Due to the scarcity of text-based open-source tools, we compare our model with leading commercial solutions (Runway, Kling-O1~\cite{Kling-O1}, and Seedance 2.0~\cite{seedance2.0}) and open-source V2V baselines (VACE~\cite{jiang2025vace} and AnyV2V~\cite{kuanyv2v}). To bridge the gap of the open-source V2V models, which support frame-conditioned generation, we utilize Qwen-Image~\cite{qwenimage} to stylize the first frame as a reference, ensuring a consistent and fair evaluation across different modalities.


Quantitative evaluations are performed across three dimensions: style quality, video quality, and human preference. 
Recent alignment research has explored adaptive preference scheduling, real-time reward modeling, and computation-aware stopping~\cite{huang2025adaptive,Huang2026RealTimeAR,huang2026does}. In line with this broader trend toward model-based assessment, we leverage Gemini-3-pro~\cite{team2023gemini} to perform automated pairwise comparisons for style similarity. Specifically, for each test case, the model is tasked with identifying which generated video aligns more closely with the reference style; the performance of each method is then quantified by its overall win rate. 
Furthermore, we assess style consistency by calculating the CSD scores~\cite{csd} between each subsequent frame and the initial frame. This metric serves to measure the stylistic stability across the temporal dimension, ensuring that the stylistic attributes remain coherent throughout the video sequence.
Following the established settings of VBench~\cite{huang2024vbench}, we assess video quality through motion smoothness and imaging quality. In addition to standard metrics, we employ Gemini-3-pro~\cite{team2023gemini} to perform a comprehensive automated assessment of the results, focusing on dynamic quality, static quality, and quality. Furthermore, our human preference evaluation focuses on both artistic scores and fine-grained content alignment with the input videos. We refer the reader to the Supplementary Material for more details.

\begin{table}[t]
\centering
\caption{Quantitative comparison between EchoStyle and SOTA methods. The \textbf{best} and \underline{second-best} results are highlighted in \textbf{bold} and \underline{underline}, respectively.}
\label{tab:comparison}

\renewcommand{\arraystretch}{1.3} 
\setlength{\arrayrulewidth}{0.4pt}
\setlength{\tabcolsep}{4pt}
\begin{adjustbox}{max width=0.95\textwidth}
\begin{tabular}{c|cc|ccccc}
\noalign{\hrule height 1pt}
& \multicolumn{2}{c|}{Style Quality} & \multicolumn{5}{c}{Video Quality} \\ \hline
Method & \makecell{Style\\Similarity} & \makecell{Style\\Consis.} & \makecell{Motion\\Smooth.} & \makecell{Imaging\\Qual.} & \makecell{Static\\Qual.} & \makecell{Dynamic\\Qual.} & \makecell{Aesthetic\\Qual.} \\ \noalign{\hrule height 1pt}
AnyV2V      & 0.351 & \underline{0.877} & 0.957 & 0.654 & 0.733 & 0.611 & 0.815 \\
VACE-Depth  &0.280 & 0.805 & 0.972 & 0.687 & 0.845 & 0.700 & 0.837 \\
VACE-Flow   &0.394 & 0.819 & 0.960 & 0.681 & 0.856 & 0.732 & 0.852 \\ \hline
Runway      & 0.239 & 0.860 & \underline{0.982} & 0.704 & 0.721 & 0.637 & 0.747 \\
Kling-O1    & \textbf{0.809} & 0.883 & \textbf{0.984} & \textbf{0.746} & \textbf{0.879} & \underline{0.775} & 0.806 \\
Seedance2.0 & 0.628 & 0.831 & 0.980 & \underline{0.725} & \underline{0.860} & 0.741 & \textbf{0.885} \\ \hline
EchoStyle   & \underline{0.802} & \textbf{0.895} & 0.966 & 0.696 & 0.850 & \textbf{0.783} & \underline{0.878} \\ \noalign{\hrule height 1pt}
\end{tabular}
\end{adjustbox}
\end{table}

\begin{figure}[!t]
    \centering
    \includegraphics[width=\textwidth]{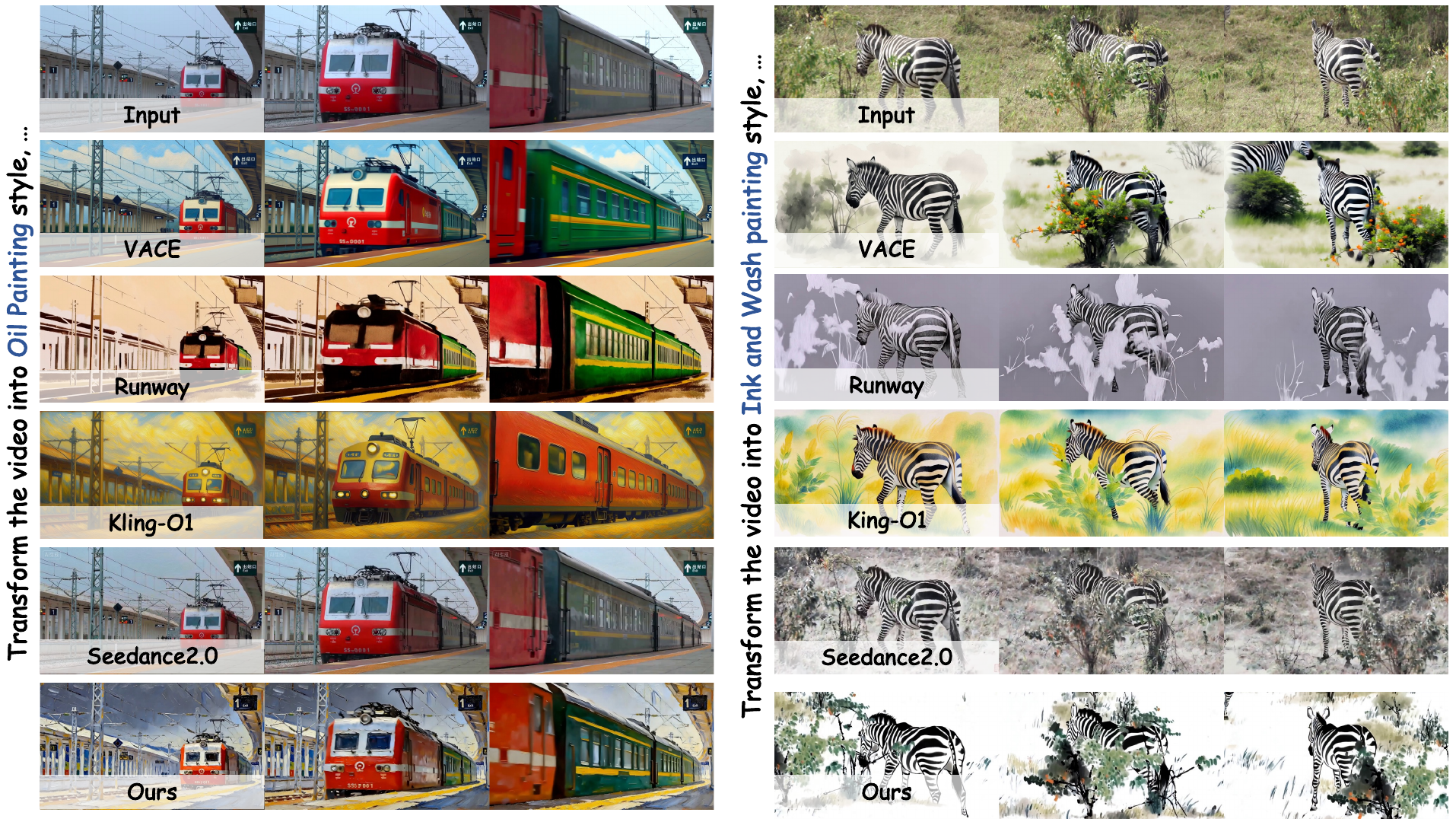} 
    \\[10pt] 
    \includegraphics[width=\textwidth]{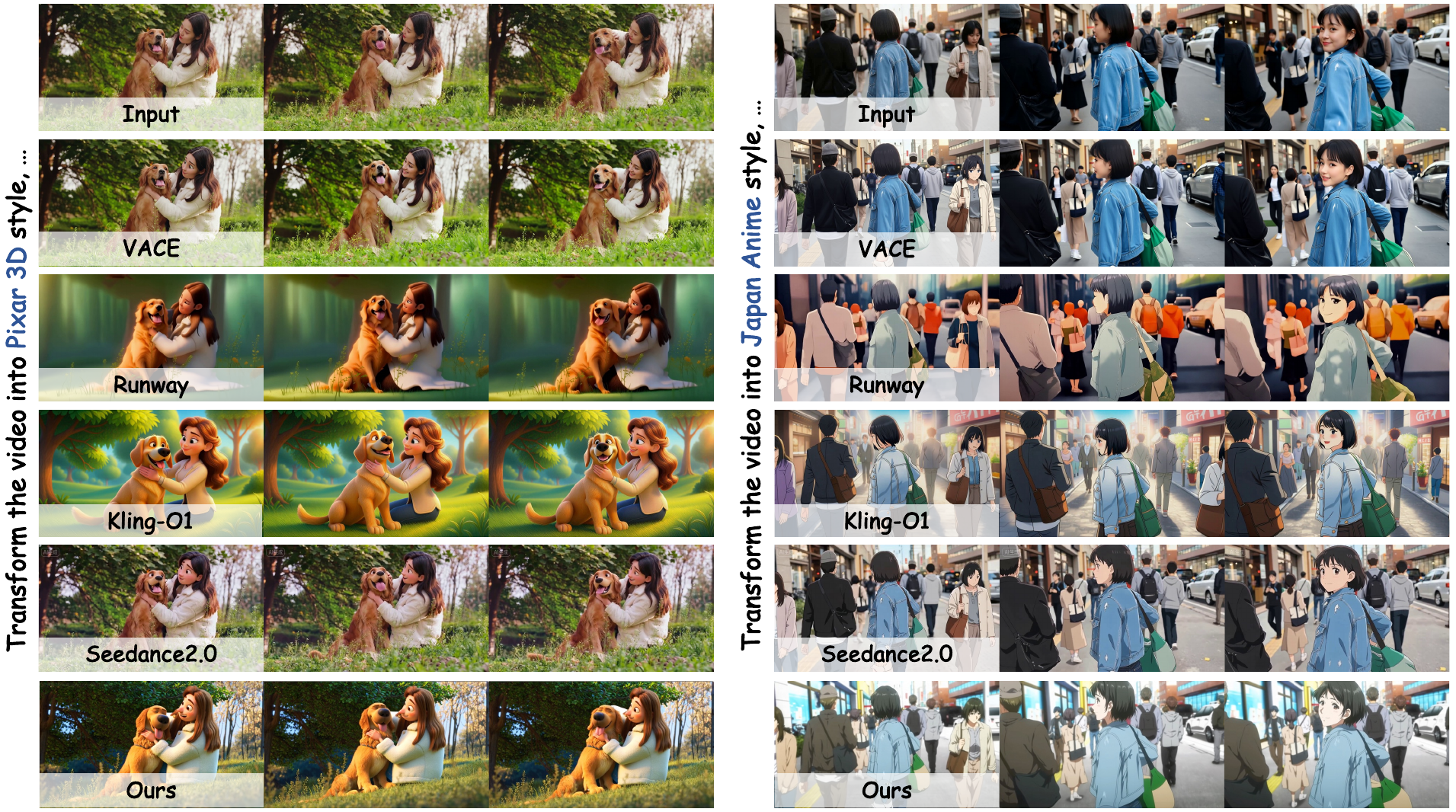}
    \caption{\textbf{Visual comparison of video stylization results.} EchoStyle demonstrates excellent visual quality and stronger temporal coherence.}
    \label{fig:compare}
\end{figure}

The qualitative and quantitative comparisons are summarized in Fig. \ref{fig:compare}, Tab.~\ref{tab:comparison} and Tab.~\ref{tab:hp}. In summary, EchoStyle demonstrates superiority over existing baselines in style quality and human preference. Regarding video quality, EchoStyle follows closely behind SOTA closed-source methods, Kling-O1 and Seedance 2.0. Significantly, we secure the top-tier ranking in dynamic quality and the second-highest score in aesthetic quality, highlighting the immense artistic utility and excellent temporal quality of our proposed method and offering a powerful alternative for professional-grade artistic video production.

\begin{wraptable}{r}{7.5cm} 
\centering
\caption{Comparison of Human Preference. The \textbf{best} and \underline{second-best} results are highlighted in \textbf{bold} and \underline{underline}, respectively.} 
\label{tab:hp}
\renewcommand{\arraystretch}{1.3} 
\small 
\begin{adjustbox}{max width=0.55\textwidth}
\begin{tabular}{cccc}
\noalign{\hrule height 1pt}
method & \makecell{Style\\ Similarity} & \makecell{Style\\ Consistency} & \makecell{Content\\ Preservation} \\ \noalign{\hrule height 1pt}
AnyV2V         & 0.490                                                      & 0.768                                                            & 0.790                                                        \\
VACE-Depth     & 0.489                                                      & 0.779                                                         & 0.850                                                     \\
VACE-Flow      & 0.472                                                      & 0.765                                                   & 0.855                                                     \\\hline
Runway         & 0.386                                                      & 0.849                                                         & 0.844                                                     \\
Kling-O1        & \underline{0.815}                                                      & \underline{0.853}                                                         & 0.833                                               \\
Seedance2.0    & 0.792                                               & 0.821                                                        & \textbf{0.889}                                                     \\ \hline

EchoStyle & \textbf{0.826}                                            & \textbf{0.870}                                                & \underline{0.860}                                            \\ \noalign{\hrule height 1pt}
\end{tabular}
\end{adjustbox}
\end{wraptable}
Although Kling-O1 and Seedance 2.0 offer high visual quality, they exhibit several drawbacks in the task of fine-grained video stylization. As shown in the bottom two subfigures of Fig.~\ref{fig:compare}, Kling-O1 exhibits limited capacity in synthesizing complex scenes, often resulting in sparse details. Its generated results exhibit monotonous color schemes and simplistic lighting hierarchies, leading to lower human-rated style similarity compared to our method. Runway, while preserving alignment, suffers from suboptimal visual quality and limited expressiveness, failing to faithfully transition to the target styles. 
Seedance 2.0 achieves the best results in terms of human preference for content preservation. However, as shown in the top two subfigures of Fig. ~\ref{fig:compare}, it often struggles to maintain \textit{high-frequency stylistic details}, such as those in oil painting and ink art styles. Furthermore, we observe that its output is prone to \textit{style decoupling} between the subject and the background, as illustrated in Fig.~\ref{fig:compare} bottom-left case.
For first-frame-guided baselines, VACE experiences significant ``style loss". 
As the video progresses, the stylized features increasingly revert to their realistic counterparts.
In the stylized output, the stylistic attributes are progressively lost, converging towards the original, non-stylized frames.
Additionally, AnyV2V is constrained by its architecture to sequences under 16 frames. In summary, EchoStyle consistently outperforms competing methods across style quality, video aesthetics, and human preference evaluation.

\begin{figure}[!t]
    \centering
    \includegraphics[width=\textwidth]{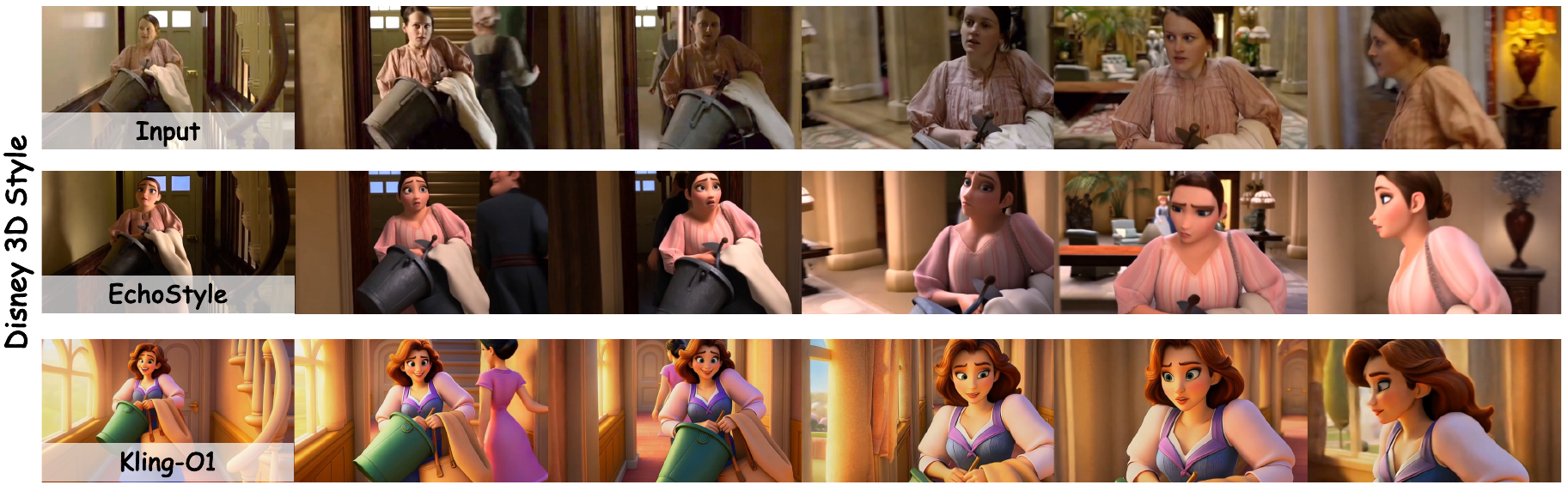} 
    \\[10pt] 
    \includegraphics[width=\textwidth]{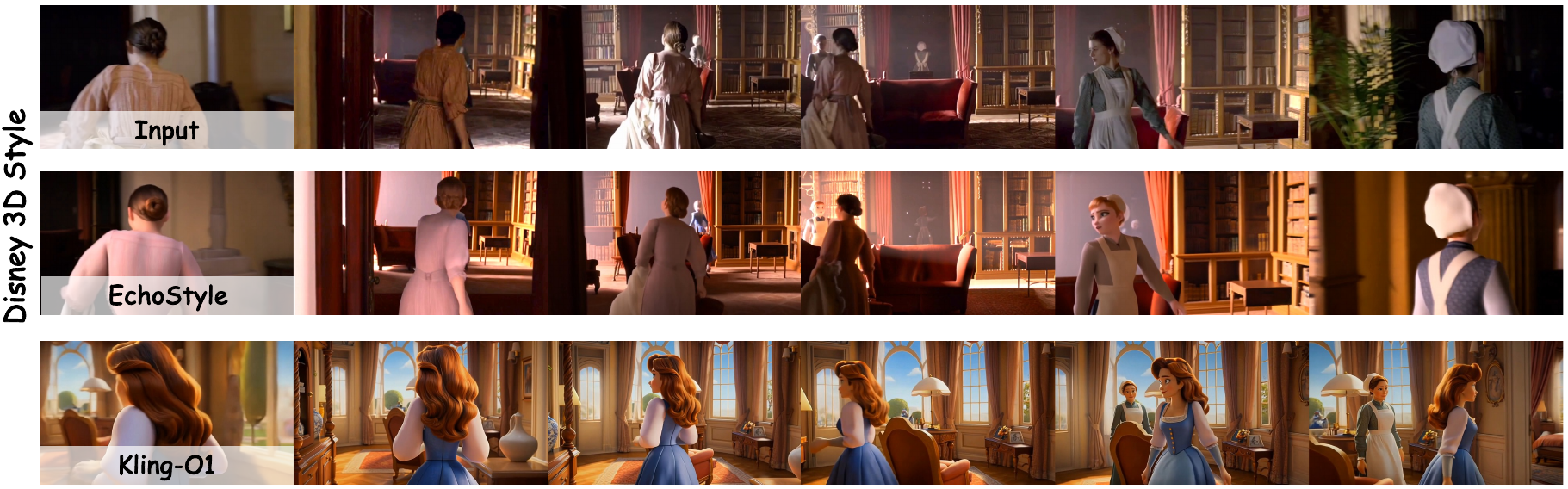}
    \caption{Visual demonstration of temporally consistent stylization on extended video sequences. Our method demonstrates superior performance in maintaining content preservation, illumination depth, and artistic texture.}
    \label{fig:long_com}
\end{figure}

\subsection{Long Video Extension}
To address the inherent challenges of content preservation and quality degradation in long-duration video generation, we systematically evaluate the robustness of our framework. 
Our experiments utilize high-quality cinematic long shots characterized by complex spatial dynamics and intricate illumination variations, which impose stringent requirements on temporal stability. 

For comparison, we employ Kling-O1 as the baseline. Specifically, for its long-duration extension, we utilize the last frame of the preceding video segment in conjunction with the reference video as conditioning prompts. The qualitative results are presented in Fig.~\ref{fig:long_com}. Both models successfully achieve the target stylistic transformation, characterized by exaggerated facial features (e.g., enlarged eyes) and smoothed skin textures typical of modern 3D animation. However, EchoStyle demonstrates superior stylistic distillation and visual refinement, while Kling-O1 tends to produce overly simplified textures and flatter lighting. At the same time, Kling-O1 occasionally suffers from semantic drift, leading to quicker error accumulation, which causes the character motions and environmental scenes in the latter stages of the video to deviate significantly from the original source content.
\subsection{Ablation Study}
\begin{figure}[t]
    \centering
    \includegraphics[width=\textwidth]{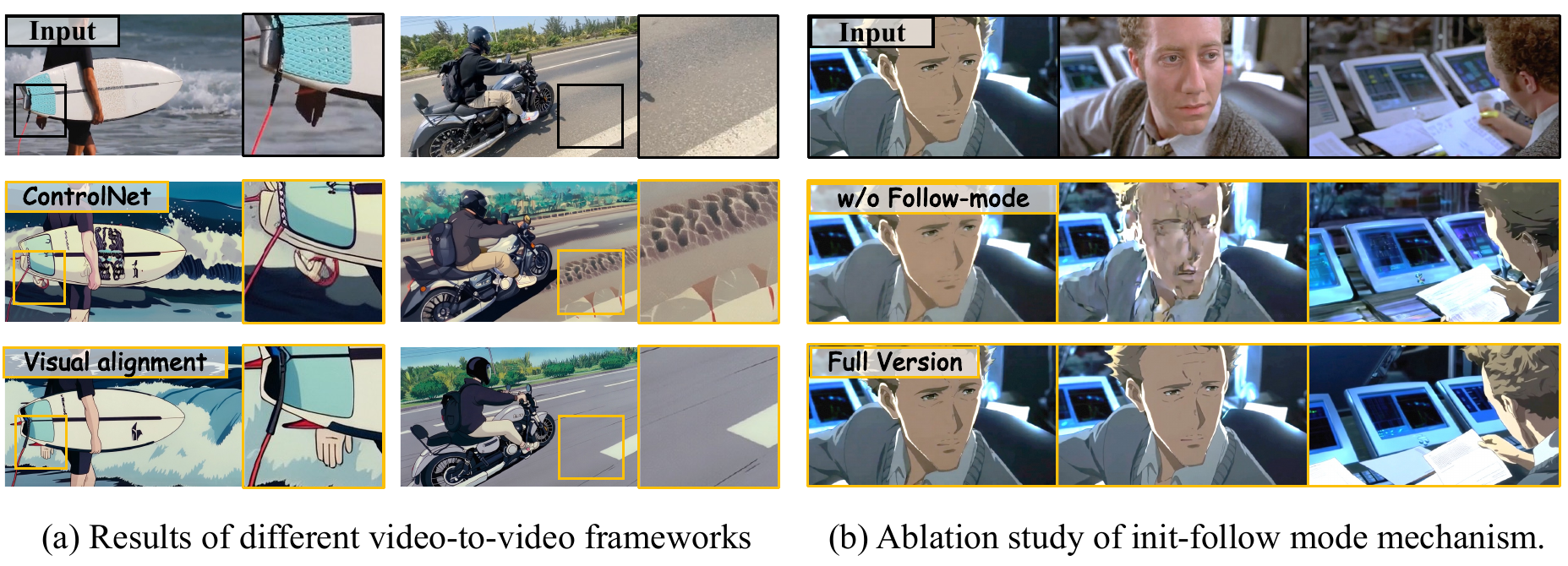}
    \caption{Qualitative ablation study of model framework and temporal conditioning.}
    \label{fig:ab}
\end{figure}


We conduct a series of ablation experiments to validate the effectiveness of the respective designs. Due to space constraints, we refer the reader to the Supplemental Material for further details and analysis.

\noindent\textbf{Effectiveness of Model Framework.} 
We compare our utilized \textit{Visual Alignment} strategy against two common baselines: \textit{ControlNet} and \textit{In-context conditioning}. 
As Fig.~\ref{fig:ab}(a) illustrates, \textit{ControlNet}~\cite{contrlnet} is prone to generating localized artifacts due to its rigid, pixel-level feature fusion mechanism. 
The \textit{In-context}~\cite{incontext} approach, which incorporates additional video frames, doubles the token sequence length, leading to significantly slower training convergence. More details of this baseline are provided in the Sup. Material. 
In contrast, our \textit{Multi-Visual-Condition Alignment} design achieves the optimal balance between generative quality and computational efficiency.


\noindent\textbf{Ablation for init-follow mode mechanism.} 
We perform an ablation study to evaluate the impact of temporal priors in follow-mode training. The visual comparisons in Fig.~\ref{fig:ab}(b) demonstrate that without explicit temporal supervision, the model suffers from \textit{temporal discontinuities} at the junction of recurrent segments. Even with a strong foundation model, the instantaneous shift in input triggers undesirable artifacts during the transition. 
Our init-follow-mode approach, which takes advantage of both temporal conditioning and mask-based temporal encoding, successfully addresses this challenge. This design enables the model to bridge the gap between recurrent segments, leading to superior temporal stability and consistency in extended sequences.

\section{Conclusion}
We propose EchoStyle, a stable and expandable framework for text-driven stylization of videos with arbitrary lengths. We design a text-driven video-to-video architecture to achieve refined stylistic distillation. Furthermore, we introduce a robust and reusable reverse data generation pipeline to construct a large-scale and high-quality video stylization dataset, \textit{V-Style20k}, which will be open-sourced to the community. To handle long videos, we propose an init-follow-mode mechanism and a sliding-window strategy, scaling video stylization to minute-level duration. Extensive experiments and analysis demonstrate that EchoStyle achieves superior dynamic style consistency as well as fine-grained content preservation for both short and long videos, even comparable to commercial closed-source solutions.


\section*{Acknowledgements}
This work was supported by the National Natural Science Foundation of China (Grant No. 62372480).

%
%
\bibliographystyle{splncs04}
\bibliography{main}

\end{document}